# OpenQA: Hybrid QA System Relying on Structured Knowledge Base as well as Non-structured Data

Gaochen Wu
Computer Science and Technology
Tsinghua University
wgc2019@mail.tsinghua.edu.cn

Bin Xu
Computer Science and Technology
Tsinghua University,
xubin@tsinghua.edu.cn

Yuxin Qin
Computer Science and Technology
Tsinghua University
tyx16@mails.tsinghua.edu.cn

Yang Liu
Computer Science and Technology
Tsinghua University

Lingyu Liu
Computer Science and Technology
Tsinghua University

Ziwei Wang
Information Science & Technology
Peking University

## ABSTRACT

Search engines based on keyword retrieval can no longer adapt to the way of information acquisition in the era of intelligent Internet of Things due to the return of keyword related Internet pages. How to quickly, accurately and effectively obtain the information needed by users from massive Internet data has become one of the key issues urgently needed to be solved. We propose an intelligent question-answering system based on structured KB and unstructured data, called OpenQA, in which users can give query questions and the model can quickly give accurate answers back to users. We integrate KBQA structured question answering based on semantic parsing and deep representation learning, and two-stage unstructured question answering based on retrieval and neural machine reading comprehension into OpenQA, and return the final answer with the highest probability through the Transformer answer selection module in OpenQA. We carry out preliminary experiments on our constructed dataset, and the experimental results prove the effectiveness of the proposed intelligent question answering system. At the same time, the core technology of each module of OpenQA platform is still in the forefront of academic hot spots, and the theoretical essence and enrichment of OpenQA will be further explored based on these academic hot spots.

## 1 Introduction

Open domain question answering has been a long-standing problem in the history of natural language processing (Robert F Simmons et al., 1963; Julian Kupiec et al., 1993; Ellen M Voorhees et al., 1999; Antoine Bordes et al., 2015; David Ferrucci et al., 2010; Stanislaw Antol et al., 2015; Makarand Tapaswi et al., 2016; Danqi Chen et al., 2017; Jinhyuk Lee et al., 2021; Jinhyuk Lee et al., 2021). The goal of open domain question answering is to build automated computer systems which can answer any type of (factoid) questions that people might ask, relied on a large collection of unstructured natural language documents, structured data (e.g., knowledge bases), semi-structured data (e.g., tables) and other modalities such as images or videos.

How to accurately and effectively obtain the knowledge needed by users from all kinds of massive Internet data has become a key and challenging topic in the era of intelligent Internet of Things, that is, when users give queries, the model can quickly provide accurate answers. On the one hand, question answering system based on structured data can obtain accurate information quickly by querying knowledge base. However, because of the limitation of incomplete knowledge bases usually covering less than 2% of knowledge, multi-entity multi-relationship complex questions with constraints and the non-independent identically distributed knowledge base, knowledge base question answering still faces great challenges. On the other hand, question answering system based on massive unstructured data is still one of the research hotspots of NLP. At the same time, how to build question-answering system based on both structured data and unstructured data is also a subject that must be studied in the future.

The complex non-independent identically distributed question answering over knowledge bases similar to GRAILQA (Gu Yu, et al. 2021) dataset is a current research frontier issue. Dense retrieval methods including dense passage retrieval and dense phras retrieval are proposed to solve unstructured data question answering and achieve the state of the art on open domain question answering (Jinhyuk Lee et al. 2021). DEEPQA (David Ferrucci et al. 2010) is one of the most representative modern question-answering systems, which featured in the 2011 TV game show JEOPARDY! has received great attention. It's a very complex system, made up of many different components, that relies on unstructured resources and structured data to generate candidate answers. Multi-modal question answering will also be one of the hot topics in the near future.

In order to obtain the accurate information required by users, we propose an intelligent question answering system OpenQA based on structured and unstructured data. In response to questions raised by users, OpenQA simultaneously starts three solving modules, including semantic parsing and deep learning parsing QA modules based on structured data, and two-stage question answering method based on retrieval and reading comprehension of unstructured data. Then, the three sequences are obtained by combining the answers obtained from the three solving modules and the question respectively, and are input through Transformer module. The representations of the three sequences are then through a linear layer and a softmax layer. Finally, the answer with the highest probability is output. To prove the effectiveness of our proposed intelligent QA model, we conduct several preliminary experiments on our constructed dataset. Experimental results show that the question answering model OpenQA can generate accurate answers to the questions raised by users.

To sum up, the contributions of this study mainly include the following three aspects: **(1)** We propose an intelligent question answering model OpenQA based on structured and unstructured data, which successfully integrates KBQA based on deep semantic parsing method and unstructured question answering based on retrieval and neural machine reading comprehension. **(2)** The experimental results prove that our proposed intelligent question answering model OpenQA can accurately answer the questions raised by users. **(3)** The intelligent question answering system OpenQA can be used as a platform for studying complex non-independent identically distributed KBQA problems in the near future, as well as for studying cutting-edge open domain QA and multi-modal QA.

## 2  Related Work

**Independently identically distributed (I.I.D.) KBQA**. Simple question answering over knowledge base (KBQA) tasks are also known as factoid question answering. Simple KBQA questions typically consist of an entity and a relationship that can be mapped to the knowledge base and then directly retrieve the answer entity or attribute value. Wenbo Zhao et al., (2019) obtained an accuracy rate of 85.4% on the dataset of SimpleQuestions (Antoine Bordes et al., 2015), which means that the simple question answering over knowledge bases has almost been solved. Complex KBQA tasks are challenging because they require multiple combinatorial reasoning abilities, (Alon Talmor et al., 2018, Yuyu Zhang et al., 2018, Michael Schlichtkrull et al., 2018, Mike Lewis et al., 2019). In order to speed up the study of complex KBQA, several complex KBQA datasets are proposed, for example, Jiaxin Shi et al. (2021) built KQA Pro, a large complex KBQA dataset with interpretable program and accurate SPARQL.

**Non-independent identically distributed (non-I.I.D.) KBQA**. Yu Gu et al. (2021) find that the existing research on KBQA is mainly conducted in the standard I.I.D. That is, the training distribution on the problems is the same as the test distribution. However, in a large-scale KBs, I.I.D. may be neither reasonable nor desirable because: (1) it is difficult to capture the true user distribution, and (2) it is very inefficient to randomly sample training samples from a large-scale space. Therefore, they propose that the KBQA model should have three inherent generalization: I.I.D, compositional and zero-shot generalization. To facilitate the development of strong generalization KBQA models, they build and publish a new large-scale, high-quality dataset of 64,331 questions, GRAILQA, and provide an evaluation setting for three levels of generalization capability. Xi Ye et al. (2021) proposed a rank-and-generate method for KBQA, that is, RNG-KBQA, which solves the problem of zero-shot coverage in complex KBQA by generation models while maintaining strong generalization ability. SOTA performance is achieved on GRAILQA dataset.

**QA over unstructured data**. Using unstructured data such as Wikipedia as knowledge sources makes the QA task face the dual challenges of large-scale open domain Q&A and machine understanding of text. To answer any question, first retrieve a few relevant articles from millions of articles, and then read the relevant articles carefully to find the answer. Danqi Chen et al. (2017) propose the DrQA system which essentially consists of two parts: (1) DOCUMENT RETRIEVER, which is used to search related articles; (2) DOCUMENT READER, which is used to extract answers from a single document or a collection of small documents. Zhilin Yang et al., (2018) propose a multi-hop reading comprehension dataset HOTPOT. The full-Wiki setup requires the QA model to answer questions based on all Wikipedia articles. This full-Wiki setup tests the performance of the system's open-domain multi-hop reasoning capability, requiring the model to be able to locate and reason about relevant facts.

**Dense retrieval QA method.** Recent studies have found that dense retrieval method shows a greater prospect than sparse retrieval method. Of these, dense phrase retrieval (the finest-grained search unit) is attractive because phrases can be used directly as the output of QA and slot filling tasks. The purpose of dense retrieval is to retrieve relevant context from a large corpus by learning queries and dense representations of text fragments. More recently, dense retrieval of articles (Kenton Lee et al., 2019; Vladimir Karpukhin et al., 2020; Lee Xiong et al., 2021) has been proved to be superior to traditional sparse retrieval methods such as TF-IDF and BM25 for a series of knowledge-intensive NLP tasks (Fabio Petroni et al., 2021), including open-domain questions (Danqi Chen et al., 2017) and knowledge-grounded dialogues (Emily Dinan et al., 2019). A natural design choice for these dense retrieval methods is the retrieval unit. For example, the Dense Paragraph Finder (DPR) (Vladimir Karpukhin et al., 2020) encodes a fixed-size text block of 100 words as the basic retrieval unit. At the other end, recent work (Minjoon Seo et al., 2019; Jinhyuk Lee et al., 2021) prove that phrases can be used as retrieval units. In particular, Lee et al., (2021) show that learning phrase intensive representations alone can achieve competitive performance in many open-domain QA and slot filling tasks. This is particularly appealing because phrases can be used directly as output without relying on an additional reader model to process paragraphs of text.

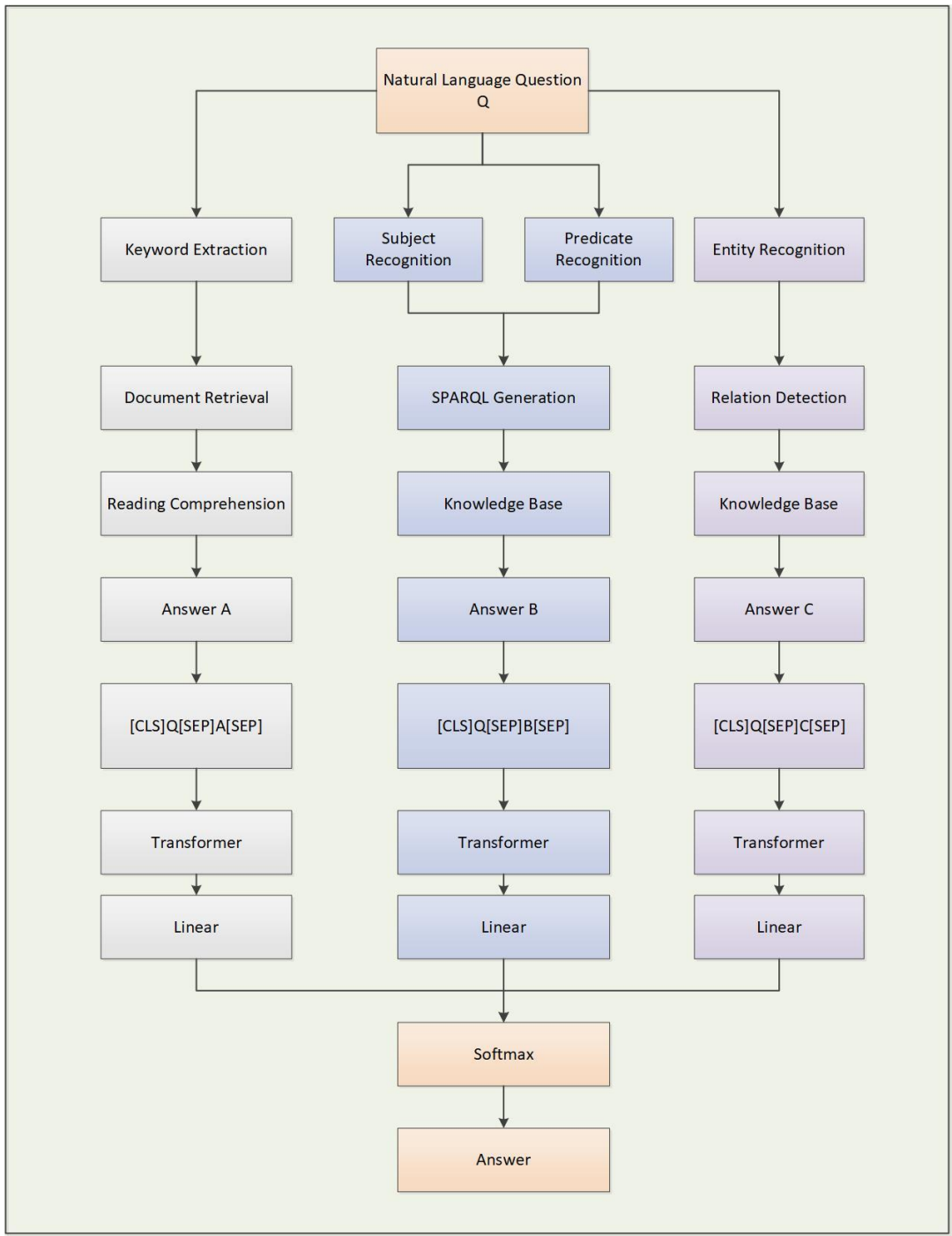

Figure 1: The Overall Architecture of OpenQA

**Multimodal QA**. IBM DEEPQA QA system (Ferrucci et al., 2010), relying on both text collections (Web pages, Wikipedia, etc.) and structured knowledge bases (FREEBASE (Kurt Bollacker et al., 2008, DBPEDIA (Soren Aueret al., 2007) resources to generate answers. YODAQA (Petr Baudis et al., 2015) is an open source system similar to DEEPQA QA system, which similarly combines various types of data resources such as websites, databases and Wikipedia. Multimodal QA task is becoming one of the most popular topics in the near future.

## 3 OpenQA

Task definition: Users input questions and The OpenQA model gives precise answers. At present, the model mainly consists of three solving modules and one answer selection module: semantic parsing module and deep learning parsing module for QA over structured data, and retrieval reading module for QA over unstructured data. The framework of the entire model is shown in **figure 1**. The following three solution modules of OpenQA will be briefly introduced, but this paper will not go into details. Please refer to our relevant papers for details.

### 3.1 Structured Data QA Solver

#### 3.1.1 Semantic Analysis QA

Task definition: Given a natural language problem (NLQ), parse out triples from NLQ and set filtering conditions. Generate SPARQL query statements and find the missing part of the triplet from the knowledge graph, that is, the answer to the question. For simple KBQA tasks, the solver is divided into the following three sub-tasks: subject recognition, predicate recognition, SPARQL query statement generation and KG query.

**Subject recognition**. In this paper, entity recognition and entity linking are carried out on the natural language questions input by users, and then a list of candidate subjects are constructed, and each subject is given a corresponding confidence according to preset rules. The entities in the knowledge graph are exported to construct the entity mapping dictionary and added into the user-defined dictionary of word segmentation tool. In the actual process, the question is segmented first, and the results are matched with the data in the entity dictionary. If they can be fully matched, they are added to the subject list as candidate subjects. By using the group capture function of regular expression template, the problem is cut and the specific capture group is added to the subject list as candidate subjects after processing.

**Predicate recognition**. Predicate recognition mainly relies on template matching. Firstly, the natural language problems are matched with the templates in the template library one by one. If the matching is successful, the predicates corresponding to the templates are added to the candidate predicates list, and each predicate is given a corresponding confidence according to the preset rules.

**SPARQL generation and query.** The subjects and predicates in the subject list and predicate list are arranged and combined one by one to generate multiple SPARQL query statements. Then the knowledge graph is queried to obtain the list of candidate answers, and the corresponding confidence is assigned to each answer according to the preset rules.

#### 3.1.2 Deep Learning Parsing QA

In this study, LD-KBQA model solver is proposed for simple KBQA problem types, such as SimpleQuestions dataset, combined with various techniques of deep learning. The model mainly consists of two core steps: entity extraction and relationship detection.

**Entity extraction.** We propose an entity extraction algorithm based on BiLSTM and Levenshtein Distance. Firstly, the BiLSTM model is used to annotate the problem statements in sequence, and the words belonging to the entity in the problem statements are obtained. Then the corresponding entity candidate set is constructed according to the result of sequence annotation. Finally, the candidate entity set is screened based on Levenshtein Distance algorithm to obtain the final entity.

**Relationship detection**. We propose a joint relationship detection algorithm based on Attentive CNN and Attentive BiGRU. Among them, Attentive CNN and Attentive BiGRU are used to extract the similarity features of sentence words and semantic similarity features in the problem respectively. Then, the similarity between the candidate relationship and the learned similarity characteristics is calculated, and the final relationship is screened according to the similarity score.

**LD-KBQA model**. We propose LD-KBQA model. LD-KBQA is implemented by combining the proposed entity extraction algorithm and relation detection algorithm. The system is used to output the corresponding answer to a given question.

### 3.2 Unstructured Data QA Solver

**Retrieve module**. This system uses Elasticsearch to build the full-text search and analysis engine. Elasticsearch is an

OpenQA

enterprise-class full text search engine with high scalability that supports concurrent access by multiple users. Before using, it is necessary to build an index for the triplets in the knowledge graph and segmented text. The index contains two fields, subject and value, corresponding to the entity in the knowledge graph and index content respectively. For the triplet data, the knowledge contained in it is first spliced, then the entity is inserted into the Subject field, and the spliced knowledge is inserted into the value field. For the segmented text, it is necessary to judge whether it can contain entities in the atlas of knowledge. If it does, the included entities should be inserted in the Subject field and segment text should be inserted in the value field.

**Reading comprehension module**. The goal of the reading comprehension model is to read the retrieved top 10 passages and extract possible answers from them. This is exactly what we set up for the span-based reading comprehension problem in the paper (Gaochen Wu et al., 2021), where the reading comprehension model can be embedded directly into OpenQA. We applied the trained reading comprehension model to the first 10 passages, which predicted an answer span with a confidence score. To make scores compatible between paragraphs in one or more retrieved documents, we use a disstandardized index and take Argmax as the final prediction for all paragraph intervals considered. It's just a very simple heuristic, and there are better ways to aggregate evidence from different paragraphs.

## 3.2 Answer Selection

The answer selection module splices the input question and the answers output by the three solution modules corresponding to the two solvers respectively as the input of Transformer (Vaswani A et al., 2017) architecture to obtain the sequence representation after splicing. The answer with the highest probability is then returned through a linear layer and a Softmax layer.

## 4 Experiments and Results

The dataset used in this study mainly comes from two sources: existing primary and secondary school teaching materials, examination papers and a large number of questions collected on the Internet. A dataset of 9020 basic education factual questions covering nine subjects are generated. The question answering pairs are divided into training set and verification set in a ratio of 7:3, and the training set are used to train the reading comprehension model. The SimpleQuestions dataset is chosen to train the LD-KBQA model. In order to verify the effectiveness of the question answering system, the accuracy rate is taken as the evaluation metric. The intelligent question answering system OpenQA achieves 85.72% accuracy on the dataset of our constructed basic education factual questions, which preliminarily verifies the validity of the method.

## 5 Conclusion

We propose an intelligent question-answering system based on structured and unstructured data, that is, OpenQA, in which users can give query questions and the model can quickly give accurate answers back to users. We integrate the KBQA structured question answering based on semantic deep learning analysis and the two-stage unstructured question answering based on machine reading comprehension into the intelligent question answering system, and return the answer with the highest probability through the answer selection module based on transformer.

## ACKNOWLEDGMENTS

OpenQA: Smart QA System Relying on Structured Knowledge Base as well as Non-structured Data.

## REFERENCES

[1] Robert F Simmons, Sheldon Klein, and Keren McConlogue. 1964. Indexing and dependency logic for answering English questions. American Documentation, 15(3):196–204.
[2] Julian Kupiec. 1993. MURAX: A robust linguistic approach for question answering using an on-line encyclopedia. In ACM SIGIR conference on Research and development in information retrieval, pages 181–190.
[3] Ellen M Voorhees. 1999. The TREC-8 question answering track report. In Text REtrieval Conference (TREC), pages 77–82.
[4] Kurt Bollacker, Colin Evans, Praveen Paritosh, Tim Sturge, and Jamie Taylor. 2008. Freebase: a collaboratively created graph database for structuring human knowledge. In Proceedings of the 2008 ACM SIGMOD international conference on Management of data, pages 1247–1250.
[5] Soren Auer, Christian Bizer, Georgi Kobilarov, Jens Lehmann, Richard Cyganiak, and ¨ Zachary Ives. 2007. DBpedia: A nucleus for a web of open data. In The Semantic Web, pages 722–735. Springer.
[6] Antoine Bordes, Nicolas Usunier, Sumit Chopra, and Jason Weston. 2015. Large-scale simple question answering with memory networks. arXiv preprint arXiv:1506.02075.
[7] David Ferrucci, Eric Brown, Jennifer Chu-Carroll, James Fan, David Gondek, Aditya A Kalyanpur, Adam Lally, J William Murdock, Eric Nyberg, John Prager, et al. 2010. Building Watson: An overview of the DeepQA project. AI magazine, 31(3):59–79.
[8] Stanislaw Antol, Aishwarya Agrawal, Jiasen Lu, Margaret Mitchell, Dhruv Batra, C Lawrence Zitnick, and Devi Parikh. 2015. VQA: Visual Question Answering. In International Conference on Computer Vision (ICCV), pages 2425–2433.
[9] Makarand Tapaswi, Yukun Zhu, Rainer Stiefelhagen, Antonio Torralba, Raquel Urtasun, and Sanja Fidler. 2016. MovieQA: Understanding stories in movies through question answering. In Conference on computer vision and pattern recognition (CVPR), pages 4631–4640.
[10] Chen et al. (2017). Reading Wikipedia to Answer Open-domain Questions. In ACL.
[11] Jinhyuk Lee, Mujeen Sung, Jaewoo Kang, and Danqi Chen. 2021. Learning dense representations of phrases at scale. In Proceedings of the 59th Annual Meeting of the Association for Computational Linguistics and the 11th International Joint Conference on Natural Language Processing (Volume 1: Long Papers), pages 6634–6647, Online. Association for Computational Linguistics.
[12] Lee, J., Wettig, A., & Chen, D. (2021). Phrase Retrieval Learns Passage Retrieval, Too. *ArXiv, abs/2109.08133*.
[13] Gu, Yu, Sue E. Kase, Michelle T. Vanni, Brian M. Sadler, Percy Liang, Xifeng Yan and Yu Su. "Beyond I.I.D.: Three Levels of Generalization for Question Answering on Knowledge Bases." *Proceedings of the Web Conference 2021* (2021): n. pag.
[14] Wenbo Zhao, Tagyoung Chung, Anuj Goyal, and Angeliki Metallinou. 2019. Simple Question Answering with Subgraph Ranking and Joint-Scoring. In NAACL-HLT
[15] Jiaxin Shi et al. (2021) KQA Pro: A Large-Scale Dataset with Interpretable Programs and Accurate SPARQLs for Complex Question Answering over Knowledge Base.
[16] Alon Talmor and Jonathan Berant. 2018. The Web as a Knowledge-Base for Answering Complex Questions. In NAACL-HLT. (ComplexWebQuestions)

[17] Yuyu Zhang, Hanjun Dai, Zornitsa Kozareva, Alexander J Smola, and Le Song. 2018. Variational reasoning for question answering with knowledge graph. In AAAI. (MetaQA)
[18] Michael Schlichtkrull, Thomas N Kipf, Peter Bloem, Rianne Van Den Berg, Ivan Titov, and Max Welling. 2018. Modeling relational data with graph convolutional networks. In ESWC.
[19] Mike Lewis, Yinhan Liu, Naman Goyal, Marjan Ghazvininejad, Abdelrahman Mohamed, Omer Levy, Ves Stoyanov, and Luke Zettlemoyer. 2019. Bart: Denoising sequence-to-sequence pre-training for natural language generation, translation, and comprehension. arXiv preprint arXiv:1910.13461 (2019).
[20] Ye, Xi, Semih Yavuz, Kazuma Hashimoto, Yingbo Zhou and Caiming Xiong. "RnG-KBQA: Generation Augmented Iterative Ranking for Knowledge Base Question Answering." *ArXiv* abs/2109.08678 (2021): n. pag.
[21] Zhilin Yang, Peng Qi, Saizheng Zhang, Yoshua Bengio, William W Cohen, Ruslan Salakhutdinov, and Christopher D Manning. 2018. Hotpotqa: A dataset for diverse, explainable multi-hop question answering. arXiv preprint arXiv:1809.09600 (2018).
[22] Kenton Lee, Ming-Wei Chang, and Kristina Toutanova. 2019. Latent retrieval for weakly supervised open domain question answering. In Proceedings of the 57th Annual Meeting of the Association for Computational Linguistics, pages 6086–6096, Florence, Italy. Association for Computational Linguistics.
[23] Vladimir Karpukhin, Barlas Oguz, Sewon Min, Patrick Lewis, Ledell Wu, Sergey Edunov, Danqi Chen, and Wen-tau Yih. 2020. Dense passage retrieval for open-domain question answering. In Proceedings of the 2020 Conference on Empirical Methods in Natural Language Processing (EMNLP), pages 6769– 6781, Online. Association for Computational Linguistics.
[24] Lee Xiong, Chenyan Xiong, Ye Li, Kwok-Fung Tang, Jialin Liu, Paul N. Bennett, Junaid Ahmed, and Arnold Overwijk. 2021. Approximate nearest neighbor negative contrastive learning for dense text retrieval. In International Conference on Learning Representations.
[25] Fabio Petroni, Aleksandra Piktus, Angela Fan, Patrick Lewis, Majid Yazdani, Nicola De Cao, James Thorne, Yacine Jernite, Vladimir Karpukhin, Jean Maillard, Vassilis Plachouras, Tim Rocktäschel, and Sebastian Riedel. 2021. KILT: a benchmark for knowledge intensive language tasks. In Proceedings of the 2021 Conference of the North American Chapter of the Association for Computational Linguistics: Human Language Technologies, pages 2523–2544, Online. Association for Computational Linguistics.
[26] Emily Dinan, Stephen Roller, Kurt Shuster, Angela Fan, Michael Auli, and Jason Weston. 2019. Wizard of wikipedia: Knowledge-powered conversational agents. In 7th International Conference on Learning Representations, ICLR 2019, New Orleans, LA, USA, May 6-9, 2019. OpenReview.net.
[27] Minjoon Seo, Jinhyuk Lee, Tom Kwiatkowski, Ankur Parikh, Ali Farhadi, and Hannaneh Hajishirzi. 2019. Real-time open-domain question answering with dense-sparse phrase index. In Proceedings of the 57th Annual Meeting of the Association for Computational Linguistics, pages 4430–4441, Florence, Italy. Association for Computational Linguistics.
[28] Jinhyuk Lee, Mujeen Sung, Jaewoo Kang, and Danqi Chen. 2021. Learning dense representations of phrases at scale. In Proceedings of the 59th Annual Meeting of the Association for Computational Linguistics and the 11th International Joint Conference on Natural Language Processing (Volume 1: Long Papers), pages 6634–6647, Online. Association for Computational Linguistics.
[29] Petr Baudis. 2015. YodaQA: a modular question answering system pipeline. In ˇ POSTER 2015—19th International Student Conference on Electrical Engineering, pages 1156– 1165.
[30] Wu, Gaochen, Bin Xu, Yuxin Qin, Fei Kong, Bangchang Liu, Hongwen Zhao and Dejie Chang. "Improving Low-resource Reading Comprehension via Cross-lingual Transposition Rethinking." *ArXiv* abs/2107.05002 (2021): n. pag.
[31] Vaswani A, Shazeer N, Parmar N, et al. Attention is all you need [C]. In Advances in neural information processing systems. 2017: 5998–6008.